\documentclass{article}

\PassOptionsToPackage{numbers, compress, sort}{natbib}

\usepackage[preprint]{neurips_2023_modified}

\usepackage[utf8]{inputenc} 
\usepackage[T1]{fontenc}    
\usepackage{hyperref}       
\usepackage{url}            
\usepackage{booktabs}       
\usepackage{amsfonts}       
\usepackage{nicefrac}       
\usepackage{microtype}      
\usepackage{xcolor}         
\hypersetup{
    colorlinks = true,
    linkbordercolor = {white},
    citecolor={olive}
}
\usepackage{multirow}
\usepackage{float}
\usepackage{amssymb}
\usepackage{array}
\usepackage{subcaption}
\usepackage{mathtools} 
\usepackage{nicematrix}
\usepackage{stmaryrd}
\usepackage{relsize}
\usepackage{colortbl}
\usepackage{nccmath}
\usepackage{array}

\usepackage{tabularray}
\usepackage{caption}
\usepackage{subcaption}

\title{Zero-Shot Uncertainty Quantification using Diffusion Probabilistic Models}

\author{%
  Dule Shu \\
  Department of Mechanical Engineering\\
  Carnegie Mellon University\\
  Pittsburgh, PA 15213 \\
  \texttt{dules@andrew.cmu.edu} \\
  \And
  Amir Barati Farimani \\
  Department of Mechanical Engineering\\
  Carnegie Mellon University \\
  Pittsburgh, PA 15213 \\
  \texttt{barati@cmu.edu}
}

\begin{document}

\maketitle
\vspace{-1mm}
\begin{abstract}
  The success of diffusion probabilistic models in generative tasks, such as text-to-image generation, has motivated the exploration of their application to regression problems commonly encountered in scientific computing and various other domains. In this context, the use of diffusion regression models for ensemble prediction is becoming a practice with increasing popularity. Under such background, we conducted a study to quantitatively evaluate the effectiveness of ensemble methods on solving different regression problems using diffusion models. We consider the ensemble prediction of a diffusion model as a means for zero-shot uncertainty quantification, since the diffusion models in our study are not trained with a loss function containing any uncertainty estimation. Through extensive experiments on 1D and 2D data, we demonstrate that ensemble methods consistently improve model prediction accuracy across various regression tasks. Notably, we observed a larger accuracy gain in auto-regressive prediction compared with point-wise prediction, and that enhancements take place in both the mean-square error and the physics-informed loss. Additionally, we reveal a statistical correlation between ensemble prediction error and ensemble variance, offering insights into balancing computational complexity with prediction accuracy and monitoring prediction confidence in practical applications where the ground truth is unknown. Our study provides a comprehensive view of the utility of diffusion ensembles, serving as a useful reference for practitioners employing diffusion models in regression problem-solving.
  
\end{abstract}

\vspace{-2mm}
\section{Introduction}


Ensemble learning \cite{kunapuli2023ensemble} has long been used to improve the accuracy of prediction by combining multiple individual machine learning models. Applying ensemble learning to deep learning models has been an active area of research \cite{lakshminarayanan2017simple, zhang2021novel, arpit2022ensemble, he2020bayesian, fort2019deep, hoffmann2021deep, el2023deep, seligmann2024beyond}, largely due to the success of deep learning in various application fields, with Deep Ensembles \cite{lakshminarayanan2017simple} being one of the representative works in this direction. Initially introduced as a simple and scalable method for estimating the predictive uncertainty of deep learning models, Deep Ensembles proposes to use a neural network architecture to predict both the mean and the variance of the target distribution, and formulates the training loss as the log-likelihood of a Gaussian distribution instead of the mean squared error of predictions. Deep Ensembles have been shown to enhance the model performance in terms of both prediction accuracy and uncertainty quantification in various applications \cite{yang2023survey, wasay2020mothernets, wu2021quantifying, li2024uncertainty, zhang2021novel, cao2020ensemble, wenzel2020hyperparameter}. Since its original formulation was introduced without a Bayesian inference form, Deep Ensembles was considered as an alternative and competing method to Bayesian neural networks \cite{neal2012bayesian, sun2020physics, franchi2023encoding} until the Bayesian formulation \cite{hoffmann2021deep} of it is proposed using Bayesian model averaging \cite{hoeting1999bayesian}. In this work, we adopt the Bayesian formulation of Deep Ensembles and generalize this formulation to the diffusion ensembles, as further specified in Section \ref{ch:ensembled-prediction}.

The development of diffusion probabilistic models can be tracked to the method for learning a distribution inspired by Non-equilibrium Thermodynamics \cite{sohl2015deep}, and has gained a fast growth in popularity and technical advancement since their success on unconditional image generation tasks with the introduction of score network \cite{song2019generative} and denoising diffusion probabilistic models (DDPM) \cite{ho2020denoising}. The outstanding performance in generating imagery data motivates the exploration of using diffusion model as a regression model in tasks such as spatio-temporal data prediction \cite{li2023seeds, price2023gencast, feng2023diffpose, mao2023leapfrog, lippe2024pde, kohl2023turbulent, gao2024bayesian, ovadia2023ditto}, super-resolution and inpainting \cite{shu2023physics, dong2024building, laroche2024fast, lugmayr2022repaint}, bioinformatics \cite{yi2024graph, liu2024predicting, zhang2024pre}, molecule property prediction \cite{duan2023diffusiondepth, kim2024diffusion, guan20233d}, material property prediction \cite{jadhav2023stressd, ogoke2024inexpensive}, optical flow estimation \cite{luo2024flowdiffuser}, image segmentation \cite{amit2021segdiff} and 3D feature reconstruction \cite{ivashechkin2023denoising}. A key step of applying diffusion probabilistic models to solving regression problems is the incorporation of conditioning information to the backward diffusion process for data generation, since a regression model typically learns a mapping from a deterministic input to an deterministic output. In a diffusion probabilistic model, such mapping is often implemented as a conditional generation of the output given the input sample. Common methods to incorporate the conditioning information includes 1. Adding the gradient of the conditioning variable likelihood to the score function and 2. Using the conditioning variable to construct an intermediate state in solving the discretized stochastic differential equation. In either case, the computation of the denoising process (formulated as denoising score matching with Langevin Dynamics or ancestral sampling) interatively inject a scaled Gaussian noise sample in model prediction (\textit{e.g.}, $z_i^M$ in Eq. 2, \cite{Song2020ScoreBasedGM}), and the randomness of the noise sample results in the variability of model prediction. At a first glance, such variability might be undesirable to a regression task since the ground truth value of the prediction target is deterministic. However, it opens up the opportunity of using a diffusion regression model to generate a probabilistic output, which can be used for ensemble prediction and uncertainty quantification (UQ) \cite{rahaman2021uncertainty, wu2021quantifying, li2024uncertainty, chan2024hyper, mouli2024using, finzi2023user}. Unlike Deep Ensembles or popular UQ methods (\textit{e.g.}, Quantile Regression \cite{kivaranovic2020adaptive, pearce2018high, tagasovska2019single}, Mean Interval Score Regression \cite{wu2021quantifying, askanazi2018comparison}) which incorporate an uncertainty metric into the model training loss, a diffusion probabilistic model does not require uncertainty estimation during model training, and allows the estimated variances of multiple prediction samples to be used as a natural choice of uncertainty metric. In this sense, we propose that a diffusion regression model is a natural tool for zero-shot uncertainty quantification.

\begin{figure}[t]
    \centering
    \includegraphics[width=\linewidth]{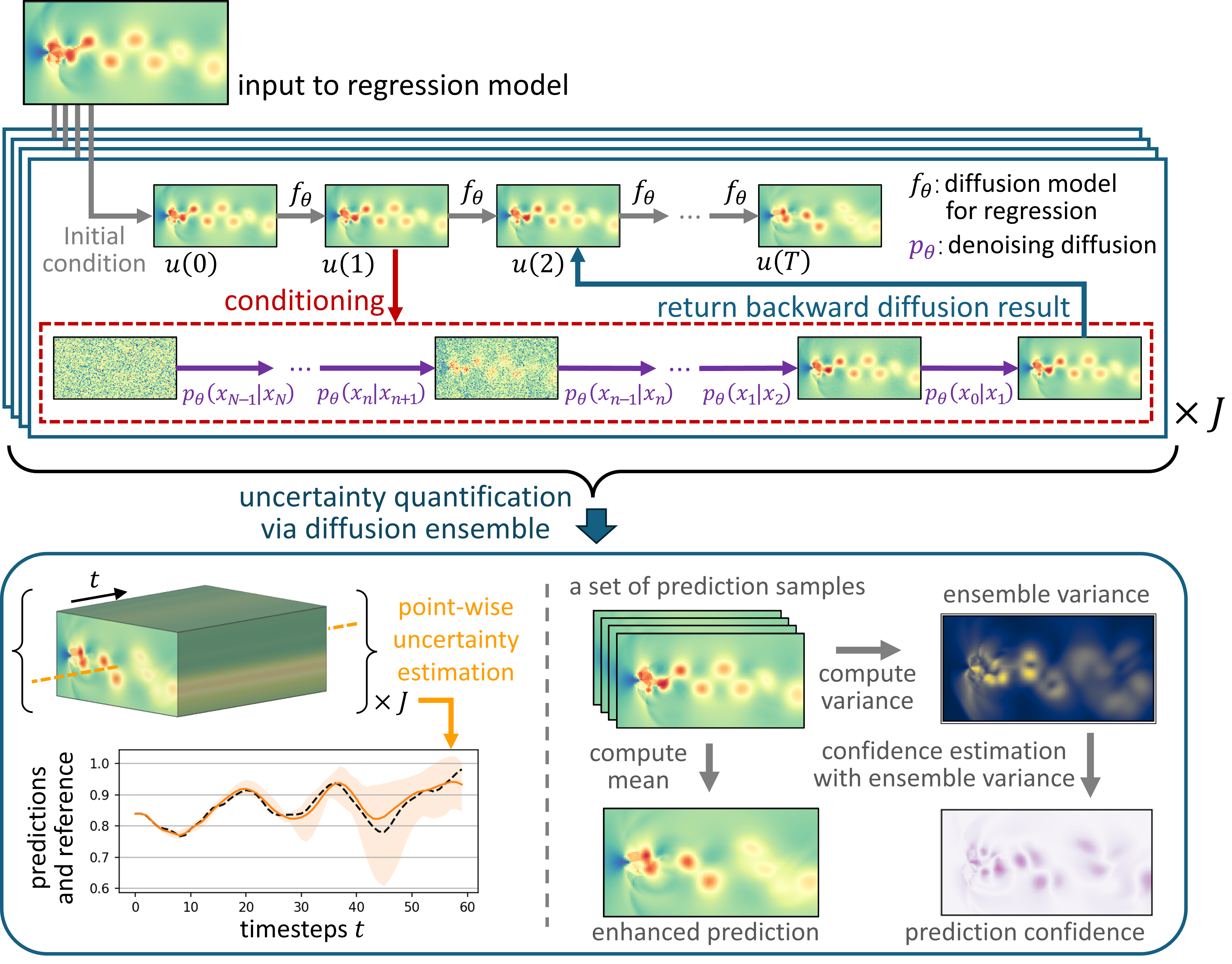}
    \caption{A overview of uncertainty quantification using diffusion probabilistic ensemble. With $J$ $(J\geq2)$ samples of model prediction, we can estimate the prediction uncertainty at a given spatial location (lower-left plot), enhance the prediction accuracy via ensembled prediction (lower-middle plot), and use the ensemble variance to evaluate the prediction confidence (lower-right plot, where regions in dark purple color indicate a lower prediction confidence).}\label{fig:method_overview}
\end{figure}

In this work, we investigate the performance of diffusion regression models for ensemble learning and uncertainty quantification. An overview of our method for utilizing diffusion models in these contexts is presented in Fig. \ref{fig:method_overview}. To enhance the generalizability of our experimental results, we selected three diffusion models with different datasets and conditioning methods. Our experiments demonstrate that ensemble consistently reduces the model prediction error across a variety of regression tasks. Additionally, we observed a high correlation between prediction error and ensemble variance, indicating that diffusion ensemble serves as a simple tool for uncertainty quantification. The contribution of this work is summarized as follows. 
\begin{enumerate}
  \item We propose a Bayesian perspective of diffusion ensemble using BMA formulation.
  \item We conduct an experiment of ensemble diffusion across different model designs and regression tasks. To the best of our knowledge, this is the first experiment of its kind to evaluate the effectiveness of ensemble on various regression diffusion models.
  \item We show through numerical experiments that diffusion ensemble can serve as a convenient tool for uncertainty quantification, due to the high correlation between ensemble variance and the ensemble prediction accuracy.
  \item We provide an analysis of the relationship between the ensemble variance and the ensemble size, offering a method to search for the appropriate ensemble size to balance computational complexity and gains in prediction accuracy.
\end{enumerate}

\section{Ensembled Prediction of Diffusion Model for Regression}
\label{ch:ensembled-prediction}
We begin with reviewing the data generation process of denoising diffusion models and then propose our interpretation of its connection to Bayesian marginalization. In the score-based generative modeling \cite{song2019generative}, the following reverse-time stochastic differential equation (SDE) is proposed to model the data generation process.
\begin{equation}
\label{eq:backward-process}
    dx = \left[f(x,t)-g(t)^2\nabla_x\log p_t (x)\right]dt + g(t)dw,
\end{equation}
where $t\in [0,T]$ is the continuous time variable, $x(0)\sim p_0$ denotes a data sample, $x(T)\sim p_T$ denotes a noisy sample with a tractable prior distribution, $f,g$ are the drift and diffusion coefficients, respectively (with respect to the forward diffusion process), $w$ denotes a standard Wiener process, and $\nabla_x\log p_t (x)$ is the score of marginal distribution at time $t$. In practice, solving Eq. \ref{eq:backward-process} to obtain $x(0)$ from $x(T)$ requires two conditions, 1. Using a neural network model $s_{\theta} (x,t)$ to estimate $\nabla_x\log p_t (x)$, and 2. Discretizing the time interval $[0,T]$ with an increasing sequence of time values $\{t_1, t_2, \cdots, t_{N-1}\}$ and applying an iteration rule $x_i = h\left(x_{i+1}\right)$ to obtain $x(0)$ from $x(T)$ (where $x_i:=x(t_i)$, $x_0:=x(0)$ and $x_N:=x(T)$). Two approaches are commonly used to achieve the second condition. The first approach, as used in DDPM sampling, computes a Markov chain $\prod_{i=1}^N p_{\theta}(x_{i-1}|x_i)$ with the following iteration rule:
\begin{equation}
\label{eq:iteration_markov}
    x_i = h(x_{i+1}, s_{\theta}):=x_{i+1}+f_1 s_{\theta} \left(x_{i+1}, t_{i+1}\right) + f_2 z_{i+1},
\end{equation}
where $f_1$,$f_2$ are functions of noise scales, and $z_{i+1}\sim \mathcal{N}\left(\mathbf{0}, \mathbf{I}\right)$. The second approach utilizes the result that for all diffusion processes modeled by an SDE, there exists a corresponding ordinary differential equation (ODE) whose trajectories share the same marginal probability densities as the SDE. Based on this derivation, the second approach either computes a deterministic iteration rule (excluding $z_{i+1}$) or employs a black-box ODE solver to solve for $x(0)$. However, Song \textit{et al.} points out that samples obtained from directly solving the probability flow ODE typically have worse FID scores \cite{song2019generative}, and subsequently introduces a predictor-corrector iteration rule incorporating a Gaussian noise sample $z\sim \mathcal{N}\left(\mathbf{0}, \mathbf{I}\right)$, which follows the same general form as shown in Eq. \ref{eq:iteration_markov}. 

Wilson \textit{et al.} \cite{wilson2020bayesian} proposes the following formulation for the predictive distribution of Bayesian model averaging (BMA).
\begin{equation*}
\label{eq:bayes_margin_ensemble}
    p\left(y|x,\mathcal{D}\right)=\int p(y|x,\textbf{w})p(\textbf{w}|\mathcal{D})d\textbf{w},
\end{equation*}
where $y$ denotes the output (\textit{e.g.}, regression values), x denotes the input, $\textbf{w}$ denotes the weights of a neural network, and $\mathcal{D}$ is the dataset. Considering the stochasticity of Eq. \ref{eq:iteration_markov} which variates samples of a backward diffusion process conditioning on $x$, one can similarly derive a non-parametric BMA form of the backward diffusion process as follows.
\begin{equation}
\label{eq:bayes_margin_diffusion}
    p\left(x_0|x,\mathcal{D}\right)=\int p(x_0|x,x_1,x_2,\cdots,x_{N})p(x_1,x_2,\cdots,x_{N}|\mathcal{D})d\mu\left(x_1, x_2, \cdots, x_{N}\right),
\end{equation}
where $x_0\equiv y$ is the target variable of regression, and $\mu$ is a probability measure of random variables $\{x_i\}_{i=1}^{N}$. The BMA form of denoising diffusion model in Eq. \ref{eq:bayes_margin_diffusion} can be estimated with a simple Monte Carlo (MC) approximation as follows.
\begin{equation}
\label{eq:ensemble_diffusion}
    p\left(x_0|x,\mathcal{D}\right)\approx \frac{1}{J} \sum_{j=1}^J p(x_0|x,\mathbf{x}_{1:N}^{(j)}),\quad \mathbf{x}_{1:N}^{(j)}\sim p(\mathbf{x}_{1:N}|\mathcal{D}),
\end{equation}
where $\mathbf{x}_{1:N}$ denotes the sequence of intermediate denoising states $\{x_i\}_{i=1}^{N}$, and $\mathbf{x}_{1:N}^{(j)}$ is the $j$th sample of $\mathbf{x}_{1:N}$. Eq. \ref{eq:ensemble_diffusion} specifies the formula for ensembled prediction of denoising diffusion models, which we refer to as diffusion ensemble in this paper.

\section{Predictive Uncertainty Estimation of Diffusion Ensemble}
We estimate the predictive uncertainty of a diffusion model for regression over a test dataset $\mathcal{D}$ consisting of $M$ data points, \textit{e.g.}, $\mathcal{D}=\{x^{(m)}, y^{(m)}\}_{m=1}^M$. For regression problems, the label $y\in\{y^{(m)}\}_{m=1}^M$ is a real-valued variable sampled from a continuous space. Given an input data sample $x\in\{x^{(m)}\}_{m=1}^M$, we use a diffusion probabilistic model $f_{\theta}$ to compute the distribution of $y$, \textit{e.g.}, $f_{\theta} (y|x)$. 
The predictive uncertainty of the model is measured by the mean and variance of prediction samples.
Following the procedure to compute ensemble prediction specified in Eq. \ref{eq:ensemble_diffusion}, for each input sample $x$, we compute $J$ prediction samples, $\{\Tilde{y}^{(j)}\}_{j=1}^J$, by recurrently sampling the Markov chain $\prod_{i=1}^N p_{\theta}(x_{i-1}|x_i)$ and assigning each sampled value of $x_0$ to $\Tilde{y}^{(j)}$. Then, the mean of model prediction can be obtained as $\Tilde{\mu}_{y}=\frac{1}{J} \sum_{j=1}^J \Tilde{y}^{(j)}$. 
Since the label $y$ is generally defined in a multi-dimensional real space (\textit{e.g.}, $y\in \mathbb{R}^d$, $d\gg 1$), to provide a straightforward scalar quantification of the prediction variance, we take the mean of the point-wise prediction variance over all $d$ dimensions as follows. 
\begin{equation}
\label{eq:average-pred-variance}
    \Bar{\sigma}^2 (y)=\frac{1}{d} \sum_{k=1}^d \Tilde{\sigma}^2(\Tilde{y}_k),\quad \Tilde{\sigma}^2(\Tilde{y}_k)=\frac{1}{J-1} \sum_{j=1}^J \left(\Tilde{y}_k^{(j)}-y_k\right)^2,
\end{equation}
where the subscript $k$ denotes the $k$th entry of the $d$-dimensional variables $y$ and $\Tilde{y}\in\{\Tilde{y}^{(j)}\}_{j=1}^J$. 

In uncertainty quantification literature, model predictive uncertainty is attributed to two types of uncertainty: aleatoric and epistemic. the aleatoric uncertainty and the epistemic uncertainty. Aleatoric uncertainty represents the intrinsic uncertainty of the prediction task and is therefore irreducible through model selection. In contrast, epistemic uncertainty represents the lack of knowledge of the model, which can be reduced by increasing the training data or improving the model. In the following section of Experiments, we show that the ensembled prediction defined by Eq. \ref{eq:ensemble_diffusion} helps reduce the prediction error attributed to the epistemic uncertainty in the regression tasks for the selected diffusion models. In general, reducing prediction error through ensemble requires that the ground truth of prediction target is sufficiently close to the mean of uncertainty estimate. In practice, users of a regression model can only assess whether their model meets this condition by testing the performance of ensemble on out-of-distribution data. A more detailed discussion on using uncertainty quantification to characterize and improve out-of-distribution learning can be found in \cite{mouli2024using}. Berry \textit{et al.} \cite{berrycasting} suggests that the epistemic uncertainty can be captured by the variance of sampled predictions of a deep ensemble model. With experimental results, we further demonstrate that the ensemble prediction variance is highly correlated with the prediction error. This observation, while intuitive due to the connection between epistemic uncertainty and prediction error, indicates that denoising diffusion models are a class of low bias model for regression tasks. Moreover, one can potentially take advantage of such correlation by using the sampled variance as a metric to assess the reliability of a diffusion model's prediction. Moreover, one can potentially leverage this correlation by using the sampled variance as a metric to assess the reliability of a diffusion model's prediction, which can be particularly helpful in practical applications where ground truth labels are unknown. Admittedly, the benefits of ensemble prediction come at the cost of increased computational complexity. Therefore, we conducted numerical experiments to show the convergence rate of the Monte Carlo (MC) approximations defined by Equations\ref{eq:ensemble_diffusion} and \ref{eq:average-pred-variance}, providing a quantified evaluation of the computational cost of ensemble prediction.

\section{Experiments}
In this section, we show the results of ensembled learning on three different diffusion models for regression, PDE-Refiner \cite{lippe2024pde}, ACDM \cite{kohl2024benchmarkingautoregressiveconditionaldiffusion}, and a physics-informed diffusion model for fluid flow super-resolution \cite{shu2023physics} (which we referred to as PI-DFS for simplicity). All three models are developed as surrogate models for simulating physical processes governed by partial differential equations. For each model, we sample multiple predictions of the target quantity of interest, compute the ensembled prediction as the estimated means by MC approximation, and compare the prediction error of the ensembled prediction and all sampled predictions. To evaluate the variance of prediction, we compute the point-wise estimated variance of the prediction samples at each location of the data domain, visualize the variance at selected locations and simulation runs, and compute the correlation between averaged variance and prediction error over simulation timesteps. In addition, we varies the value of $J$ from Equations \ref{eq:ensemble_diffusion} and \ref{eq:average-pred-variance} to show the rate of convergence of MC approximation.

\subsection{PDE-Refiner}
PDE-Refiner is a model designed to produce more accurate and stable predictions of time-dependent PDE solutions. To achieve this goal, the model learns to refine an initial prediction by applying a backward diffusion process conditioned on that initial condition. The authors suggest that the refinement process in a diffusion style helps to preserve non-dominant high spatial frequency information in PDE solutions and therefore reduces the prediction error in long time rollout (autoregressive prediction) in their numerical experiments. More specifically, given a PDE solution at a previous timestep, $u(t-\Delta t)$, PDE-Refiner uses a neural operator model $f_{\theta}$ to make an initial prediction of the PDE solution at the current step, \textit{e.g.}, $\Hat{u}^0 (t)=f_{\theta}(u(t-\Delta t))$. Next, a $K$-step iterative refinement process is applied to the initial prediction to obtain a refined prediction as follows. 
\begin{gather*}
\label{eq:pde-refiner}
    \Hat{z}=f_{\theta} (\Hat{u}^{k-1}(t) + \sigma_k z;u(t-\Delta t),k),\\
    \Hat{u}^{k} (t)=\Hat{u}^{k-1}(t) + \sigma_k z - \sigma_k \Hat{z},\quad k\in\{1,2,\cdots,K\},
\end{gather*}
where $\sigma_k$ denotes the noise scale, and $z\sim \mathcal{N} (\mathbf{0}, \mathbf{I})$. While the network architecture for $f_{\theta}$ is chosen as a U-Net \cite{gupta2022towards}, the refine-by-denoising idea of PDE-Refiner is applicable to other network architectures such as the popular Transformer family of neural PDE solvers \cite{li2024scalable, shih2024transformers, lorsung2024physics, li2022transformer, zhao2023pinnsformer, hemmasian2024pretraining, zhou2024strategies, hemmasian2024multi, li2024cafa}.

\subsection{ACDM}
Similar to PDE-Refiner, the Autoregressive Conditional Diffusion Model (ACDM) is another model for simulating turbulent flows where diffusion processes are used to enhance the temporal stability of long rollouts. Unlike PDE-Refiner, ACDM computes a complete backward diffusion process starting from noise samples to predict the physical quantity of interest at the next timestep, where the values of the physical quantity at the current timestep and other PDE-related information are incorporated via forward process into every denoising step. ACDM outperforms PDE-Refiner on benchmark experiments at a cost of lower inference speed \cite{kohl2024benchmarkingautoregressiveconditionaldiffusion}. Let $u(t)$ denote the physical quantity of interest evaluated at timesteps $t$, and let $c_f$ be the set of coefficients related to the PDE model of the simulation (\textit{e.g.}, Reynolds number or Mach number for fluid dynamics simulation). ACDM predicts $u(t)$, a state variable at current timestep, using the previous $k$ steps of the states, $\{u(t-1), u(t-2), ..., u(t-k)\}$ and the coefficients $c_f$. A basic DDPM algorithm is used to compute an $R$-step backward diffusion process starting from $x_R :=(d_R, c_R)$ and ending at $x_0 =(d_0, c_0)$, where $d_R \sim \mathcal{N}(\mathbf{0}, \mathbf{I})$, $c_0 = \{u(t-1), u(t-2), ..., u(t-k), c_f\}$, $\{c_r\}_{r=1}^R$ is obtained from $c_0$ via reparameterization, the state transition is computed by a neural network via $x_{r-1} \sim p_{\theta}(x_{r-1}|x_r)$, and $d_0$ is the final prediction of the target $u(t)$.

\subsection{PI-DFS}
To evaluate the effectiveness of ensembled prediction on a wider range of regression problems for a diffusion model, we include experimental results with the PI-DFS model. PI-DFS model is chosen because it differs from PDE-Refiner and ACDM on two key aspects. 1. Unlike PDE-Refiner or ACDM, which are used for auto-regressive prediction of time-series data given an initial condition, PI-DFS is employed for making point-wise predictions from input to output. 2. PI-DFS is designed to minimize not only the point-wise prediction error (\textit{e.g.}, the $\mathcal{L}^2$ distance between the prediction and the ground truth) but also the physics-informed loss defined as the PDE residual of model predictions. Let $u_l$, $u_h$ be a low fidelity and a high fidelity data samples of some physical quantity of interest (\textit{e.g.}, the vorticity of fluid evaluated on a 2D mesh). In model inference stage, the PI-DFS model takes $u_l$ as input and predicts the value of $u_h$. To achieve this goal, PI-DFS first uses $u_l$ to construct an estimated intermediate state of a backward diffusion process for predicting $u_h$ as follows.
\begin{gather*}
\label{eq:dfsr_starting_state}
    x_{\tau}:=\sqrt{\Bar{\alpha}_{\tau}}u_l+\sqrt{1-\Bar{\alpha}_{\tau}}\epsilon_\tau,
\end{gather*}
where $\epsilon_{\tau}\sim\mathcal{N}(\textbf{0}, \textbf{I})$, and $\Bar{\alpha}_{\tau}$ is a noise scheduling parameter at denoising timestep $\tau$ following the notation in \cite{ho2020denoising}. Then, a denoising diffusion implicit model (DDIM) \cite{song2020denoising} starting from $x_{\tau}$ is computed to obtain $x_0$, the final state of the reverse process, as a prediction of $u_h$. At each step of denoising, the corresponding state variable, denoted as $x_{\tau_i}$, is substituted into the governing PDE (\textit{e.g.}, a 2D Navier-Stokes equation) to compute the residual. The relative $\mathcal{L}^2$ norm of the residual, defined as the physics-informed residual loss, is used as classifier-free guidance \cite{ho2022classifier} to guide the data generation process for a minimized residual loss.

\subsection{Common observations from listed tasks}
In this subsection, we summarize the phenomena commonly observed in the three numerical experiments using different diffusion models presented above. This serves to analyze the general properties of diffusion models used as a deep ensemble model for regression problems. Our observations are as follows. 
1. Ensemble improves the accuracy of model predictions. 2. The accuracy of ensembled prediction is highly correlated with the variance of the prediction samples. Further comments and quantitative evaluations regarding these two observations are presented as follows.

\begin{table}[h!]
\centering
\caption{\small A comparison of accuracy between sampled predictions and ensemble predictions, where the accuracy is evaluated by the relative $\mathcal{L}^2$ error for all models and by the PDE residual loss for the PI-DFS model. For the tasks of auto-regressive predictions, the prediction error is evaluated at the final time-step. For the task of point-wise prediction, the metrics are evaluated across initial conditions and timesteps of the reference simulation data used as the test dataset.}\vspace{2mm}
\begin{tblr}{
|p{0.13\linewidth}|p{0.11\linewidth}|p{0.11\linewidth}|p{0.11\linewidth}|p{0.11\linewidth}|p{0.23\linewidth}|}
    \hline
    \SetCell[r=2]{} model & \SetCell[c=2]{} \centering prediction error & {} & \SetCell[c=2]{} \centering residual loss & {} & \SetCell[r=2]{} Task description \\ \hline 
    {} & \centering sampled mean &  \centering ensemble & \centering sampled mean & \centering ensemble & {} \\
    \hline
    \SetCell[r=3]{} PDE-Refiner & & & & & \SetCell[r=3]{} Auto-regressive \quad\quad prediction on 1D data. \\
     & 0.1192 & \textbf{0.0986} & \centering$-$ & \centering$-$ & \\
     & & & & & \\ \hline
    \SetCell[r=3]{} ACDM & & & & & \SetCell[r=3]{} Auto-regressive \quad\quad prediction on 2D data with 4 data channels. \\
     & 0.5516 & \textbf{0.3966} & \centering$-$ & \centering$-$ & \\
     & & & & & \\ \hline
    \SetCell[r=3]{} PI-DFS & & & & & \SetCell[r=3]{} Point-wise prediction on 2D data. \\
     & 0.3752 & \textbf{0.3657} & 0.2586 & \textbf{0.1846} & \\
     & & & & & \\ \hline
\end{tblr}
\label{tab:pred_error_and_residual_loss}
\end{table}

\textbf{Improvements on prediction accuracy}. The improvements on prediction accuracy across different regression tasks are shown in Figures \ref{fig:pred_error_pderefiner}, \ref{fig:pred_error_acdm}, \ref{fig:pred_error_dfsr}, and Table \ref{tab:pred_error_and_residual_loss}. Among the three tasks evaluated in this paper, auto-regressive prediction gains a larger accuracy increase than point-wise prediction (by which we mean the model inputs are always sampled from test datasets rather than from its own outputs) in terms of the relative $\mathcal{L}^2$ error. This observation implies that ensemble is an effective strategy for countering the accumulative error in auto-regression. The experimental results with PI-DFS demonstrate that ensemble is capable of reducing both the $\mathcal{L}^p$ loss and the physics-informed loss. Note that both losses are computed by averaging the corresponding metric function (\textit{e.g.}, the $\mathcal{L}^p$ distance and the PDE residual) over the spatial domain of interest. In comparison, accuracy improvement by ensemble is not observed in statistical metrics such as the kinetic energy spectrum and the vorticity distribution. It would be interesting to see whether ensemble can also improve the frequency domain accuracy metrics of model predictions such as the kinetic energy spectrum by applying to frequency-domain features (\textit{e.g.}, the Fourier coefficient features obtained from Fast Fourier Transform as proposed in \cite{li2021fourier}) in model forward propagation. 

\begin{figure}[t]
    \centering
    \includegraphics[width=0.9\linewidth]{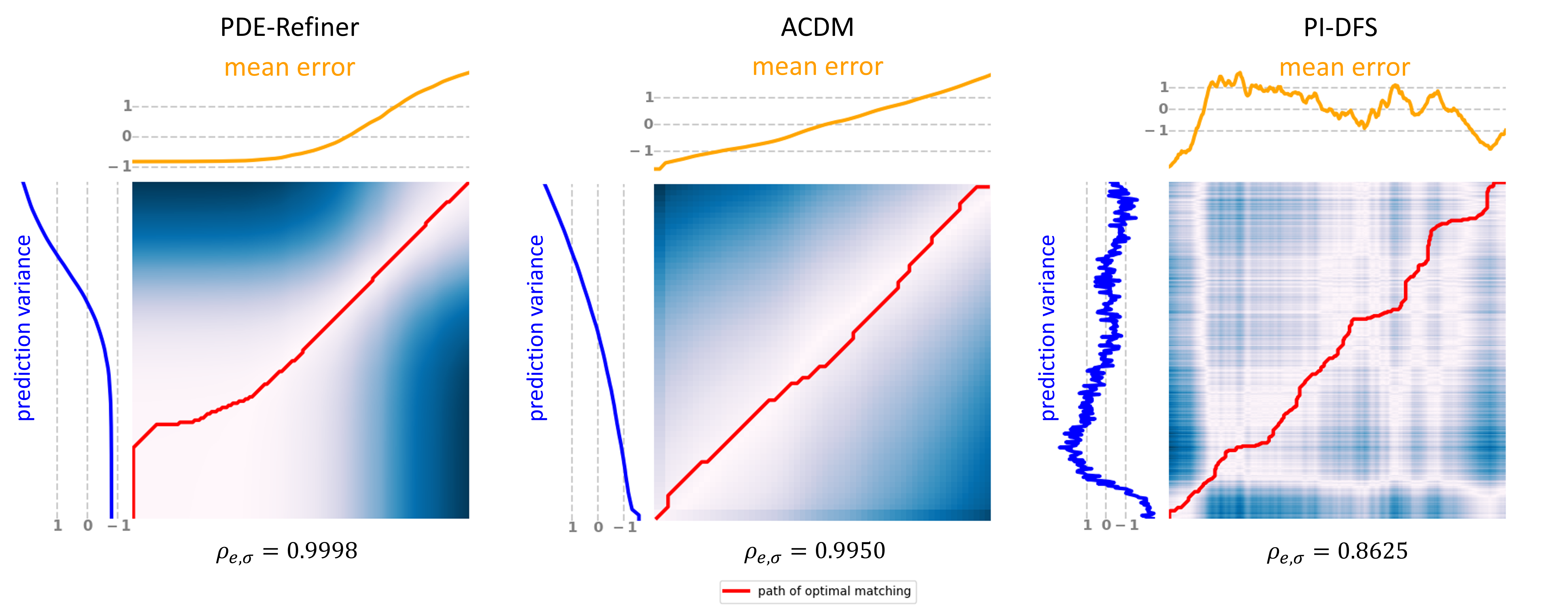}
    \caption{Dynamic Time Warping similarity between the spatial means of ensemble prediction variance and ensemble prediction error, where the Euclidean distance is used as the distance metric and the color map (varying from white to blue) for coloring the background.}\label{fig:DTW_simlarity}
\end{figure}

\textbf{Correlations between prediction variance and accuracy}. Throughout the regression tasks solved with different diffusion models in our study, we have observed a consistent correlation between the variance of the ensemble and its prediction accuracy. Specifically, it is observed that the relative $\mathcal{L}^2$ prediction error $e(t)$, defined by Eq. \ref{eq:pred_error_acdm}, is highly correlated with the ensemble variance defined as follows.
\begin{equation*}
    \begin{aligned}
        {}&\sigma_u^2(t)=\frac{1}{B\cdot D\cdot C}\sum_{b=1}^{B}\sum_{d=1}^{D}\sum_{c=1}^{C}\sigma_u^2(b,t,d,c) \\
        \text{where}\quad &\sigma_u^2(b,t,d,c) = \frac{1}{16}\sum_{j=1}^{16}\left( \Tilde{u}^{(j)} (b,t,d,c)-\mu_u (b,t,d,c) \right)^2.
    \end{aligned}
\end{equation*}
Figure \ref{fig:DTW_simlarity} shows the values of Pearson correlation coefficient (denoted as $\rho_{e,\sigma}$) as a way to evaluate the correlation between the time sequences $\{e(t)\}_{t=1}^T$ and $\{\sigma_u^2(t)\}_{t=1}^T$ \ in different tasks. Also shown in the figure are the Dynamic Time Warping (DTW) similarity between $\{e(t)\}_{t=1}^T$ and $\{\sigma_u^2(t)\}_{t=1}^T$. The results that $\rho_{e,\sigma}$ has a value close to 1 and that the optimal matching paths (marked by red solid lines) are close to the diagonal line demonstrate the high correlation between the ensemble variance and the ensemble prediction error. With such correlation, users of a diffusion model can compute the ensemble variance to assess the accuracy and reliability of the model prediction in a practical application scenario where the ground truth reference is unknown and the knowledge of the model prediction error are unavailable.

The advantage of using a diffusion model for ensembled prediction, including improvements on prediction accuracy and the utility of assessing prediction accuracy, come at a cost of increased computational complexity. To provide a quantified analysis on the trade-off between the benefit of ensemble and its computational cost, we propose the following method to examine how the ensemble variance (which are highly correlated to the prediction accuracy, as shown in Fig. \ref{fig:DTW_simlarity} ) changes with respect to the ensemble size. Let $\{\Tilde{u}^{(j)}\}_{j=1}^{N}$ be $N$ prediction samples collected for a maximal ensemble size, and let $k\in\{2,3,...,N\}$ be a chosen size of an ensembled prediction. For each $k$-combination of the $N$ samples represented by the corresponding index set $\mathcal{J}_k^{(m)}$, we compute the point-wise standard deviation of the $k$ samples as follows.
\begin{gather*}
    \sigma_k^{(m)}(b,t,d,c)=\left(\frac{1}{k}\sum_{j\in\mathcal{J}_k^{(m)}}\left(\Tilde{u}^{(j)}(b,t,d,c)-\mu_k^{(m)}(b,t,d,c)\right)^2\right)^{\frac{1}{2}} \\
    \text{where}\quad \mu_k^{(m)}(b,t,d,c) = \frac{1}{k}\sum_{j\in\mathcal{J}_k^{(m)}}\Tilde{u}^{(j)}(b,t,d,c).
\end{gather*}
Then, we compute the mean of $\sigma_k^{(m)}(b,t,d,c)$ over all data dimensions (\textit{e.g.}, $[1,B]\times[1,T]\times[1,D]\times[1,C]$), denoted as $\Bar{\sigma}_k^{(m)}$, as a scalar metric to quantify the ensemble variance for the sample index set $\mathcal{J}_k^{(m)}$. For each ensemble size $k$, we obtain a set of ensemble variance values using our proposed metric, \textit{e.g.}, $\{\Bar{\sigma}_k^{(1)}, \Bar{\sigma}_k^{(2)}, \cdots, \Bar{\sigma}_k^{(M)}\}$ where $M=C_k^N$. 
\begin{figure}[h]
    \centering
    \includegraphics[width=0.9\linewidth]{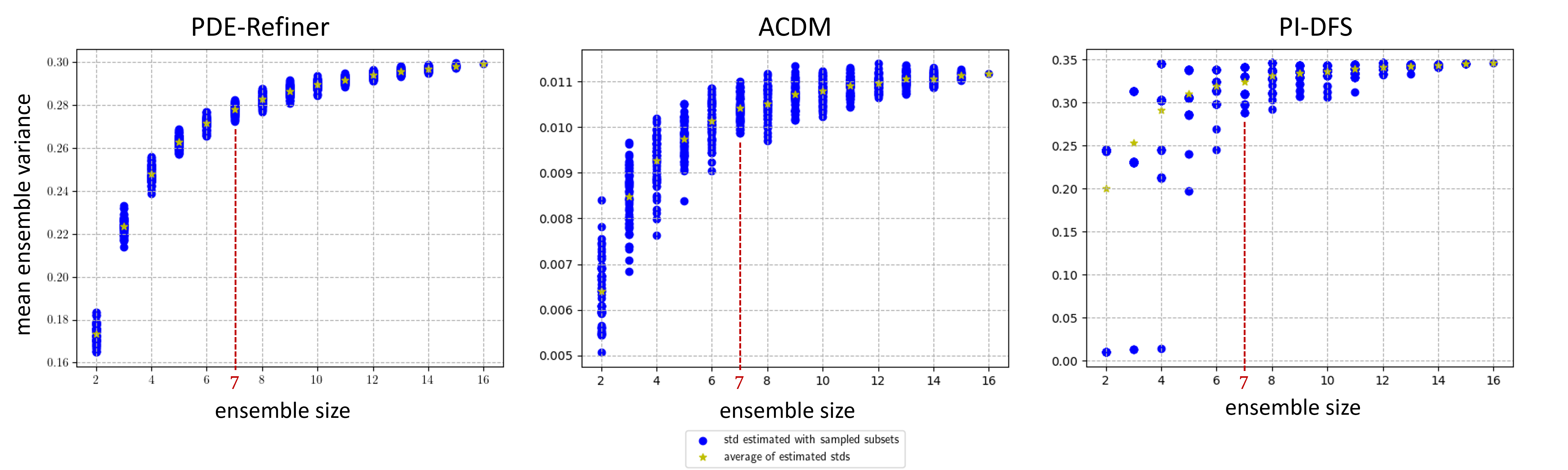}
    \caption{The plots of mean ensemble variances for different ensemble sizes. The ensemble size are chosen as 7 by observing the convergence of the mean variance with respect to the increase of ensemble size to balance the benefit of ensemble and its computational complexity.}\label{fig:ens_var_vs_ens_size}
\end{figure}
By plotting the values in the set and their mean for all ensemble sizes $2\leq k\leq N$, we can visualize how the ensemble variance converges with respect to the ensemble size, and choose a proper ensemble size for a reduced computational cost. Figure \ref{fig:ens_var_vs_ens_size} shows the plot of mean ensemble variances for different ensemble sizes. In all three tasks, a pattern of convergence of the mean ensemble variance can be observed as the ensemble size increases. As an example, we empirically choose $k=7$ by observing the plots to balance the gain of ensemble prediction with its computational complexity. With a different diffusion model, the user can adopt the same procedure to determine the size of ensemble.

\section{Conclusion}
In this work, we investigate the effect of ensemble prediction on diffusion models for regression tasks. We start with a Bayesian formulation of the ensemble prediction by incorporating the backward diffusion process into a BMA framework, then propose to use the spatial mean of point-wise ensemble variance as a metric to quantify the predictive uncertainty estimation. Numerical experiments on three different regression tasks demonstrate the effectiveness of ensemble methods on diffusion probabilistic models. Our results show that ensemble consistently improve the prediction accuracy of diffusion models, albeit to varying degrees across different tasks. Additionally, we observe a high correlation between ensemble prediction error and mean ensemble variance, indicating that diffusion ensembles provide a straightforward and convenient tool for uncertainty quantification without requiring specialized model training procedures to minimize a predefined uncertainty quantification metric. Our future work will focus on further exploring the correlation between ensemble error and variance. A promising direction is uncertainty-based importance sampling for data-efficient fine-tuning of diffusion regression models. Specifically, by modifying the model training procedure as proposed by Katharopoulos \textit{et al.} \cite{katharopoulos2018not}, the mean ensemble variance can be computed as an importance score for weighted data sampling. This approach ensures that input data samples leading to higher prediction uncertainty (and consequently higher prediction error) are more likely to be selected for querying the ground truth label, thereby minimizing data usage for model fine-tuning.

\section{Acknowledgement}
This work is supported by funding from the Division of Chemical, Bioengineering, Environmental and Transport Systems at National Science Foundation (CBET–1953222), United States. The authors would also like to thank Hongshuo Huang for helping with data sampling and Zijie Li for the inspiring discussions on problem formulation and references.

\bibliographystyle{plainnat}
\bibliography{ref}
\newpage

\begin{appendix}
\section{Experiment Details}

\subsection{PDE-Refiner}
\begin{figure}[t!]
\centering
    \begin{subfigure}{0.31\textwidth}
    \centering
        \includegraphics[width=\linewidth,height=0.95\linewidth]{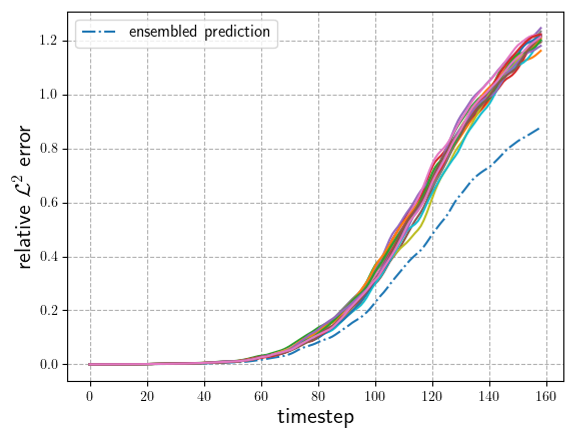}
        \caption{\small Prediction error vs. timestep.\\}
        \label{fig:pred_error_pderefiner}
    \end{subfigure}
    \begin{subfigure}{0.31\textwidth}
    \centering
        \includegraphics[width=\linewidth]{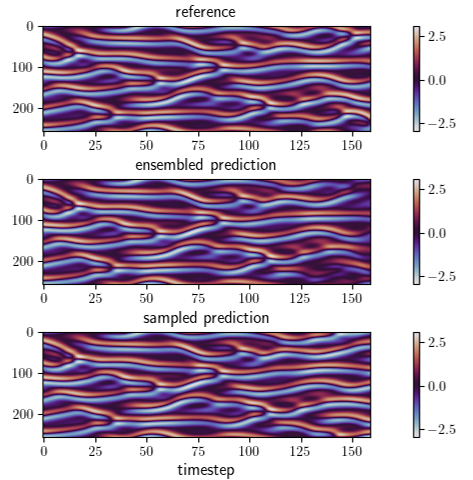}
        \caption{\small Data samples.}
        \label{fig:pred_compare_pderefiner}
    \end{subfigure}
    \begin{subfigure}{0.31\textwidth}
    \centering
        \includegraphics[width=\linewidth]{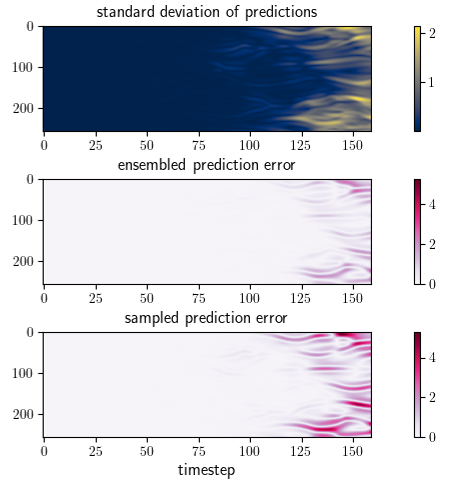}
        \caption{\small Prediction errors and variance.}
        \label{fig:pred_std_and_error_pderefiner}
    \end{subfigure}
    \caption{A comparison of prediction samples and their ensemble from PDE-Refiner.}
\end{figure}

We choose to evaluate the ensembled prediction of PDE-Refiner on the Kuramoto-Sivashinsky 1D dataset. Model training and configuration follow the procedure specified in \cite{lippe2024pde}, with a U-Net architecture conditioned on the values of timestep size, spatial resolution, and viscosity value. To collect the model prediction results, we compute roll-out on all 128 trajectories of the test dataset for a duration of 159 timesteps. For each roll-out, we collect 16 prediction samples, denoted as $\{\Tilde{u}^{(j)}\}_{j=1}^{16}$. Each sample $\Tilde{u}^{(j)}$ has a shape of $[B, T, D]$, where $B=128$ is the number of simulation trajectories, $T=159$ is the length of roll-out, and $D=256$ is the spatial dimension. The ensembled prediction is computed as $\mu_u = \frac{1}{16}\sum_{j=0}^{16} \Tilde{u}^{(j)}$. 
To compare the sampled predictions and their ensemble, we compute their relative $\mathcal{L}^2$ prediction error as follows.
\begin{gather*}
\label{eq:pde-refiner}
    e(t)=\frac{1}{B}\sum_{b=1}^{B}\frac{\|\Hat{u}(b,t)-u(b,t)\|_2}{\|u(b,t)\|_2},
\end{gather*}
where $\Hat{u}$ is either a sampled prediction or the ensembled prediction, and $\|\cdot\|_2$ denotes the $\mathcal{L}^2$-norm. A comparison of the prediction errors is shown in Fig. \ref{fig:pred_error_pderefiner}. The error of ensembled prediction increases significantly more slowly over time than the error of any sampled prediction. Figures \ref{fig:pred_error_pderefiner} and \ref{fig:pred_std_and_error_pderefiner} provide visualization of prediction for a sampled data sequence. In particualr, Fig. \ref{fig:pred_error_pderefiner} shows the rollouts of a single model prediction sample (lower subplot), the ensembled prediction (central subplot) and the reference ground truth (upper subplot). Fig. \ref{fig:pred_std_and_error_pderefiner} shows the point-wise absolute prediction error of a sampled rollout (lower subplot) and the ensembled rollout (mean: central subplot, variance: upper subplot), respectively. A quantitative comparison of model prediction errors is shown in the first row of Table \ref{tab:pred_error_and_residual_loss}, where the ensembled prediction has a lower prediction error than the average error of the 16 sampled rollouts at the final timestep. In fact, as shown in \ref{fig:pred_error_pderefiner}, the ensembled prediction starts to outperforms any sampled prediction after halfway of the simulation and maintains an increasing performance margin till the end of the simulation.


\subsection{ACDM}
\begin{figure}[h]
    \centering
    \begin{minipage}{0.6\linewidth}
        \centering
        \begin{subfigure}{\linewidth}
        \includegraphics[width=\textwidth]{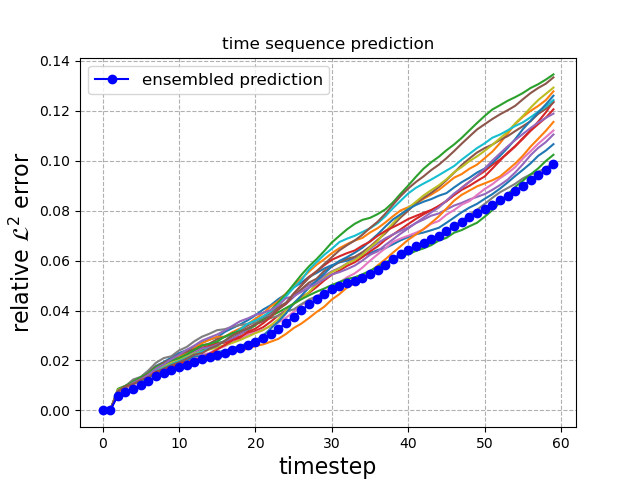}
        \caption{\small Prediction error vs. timestep.}
        \label{fig:pred_error_acdm}
        \end{subfigure}
    \end{minipage}
    \begin{minipage}{0.9\linewidth}
        \vspace{3mm}
        \begin{subfigure}{\linewidth}
        \includegraphics[width=1.0\linewidth]{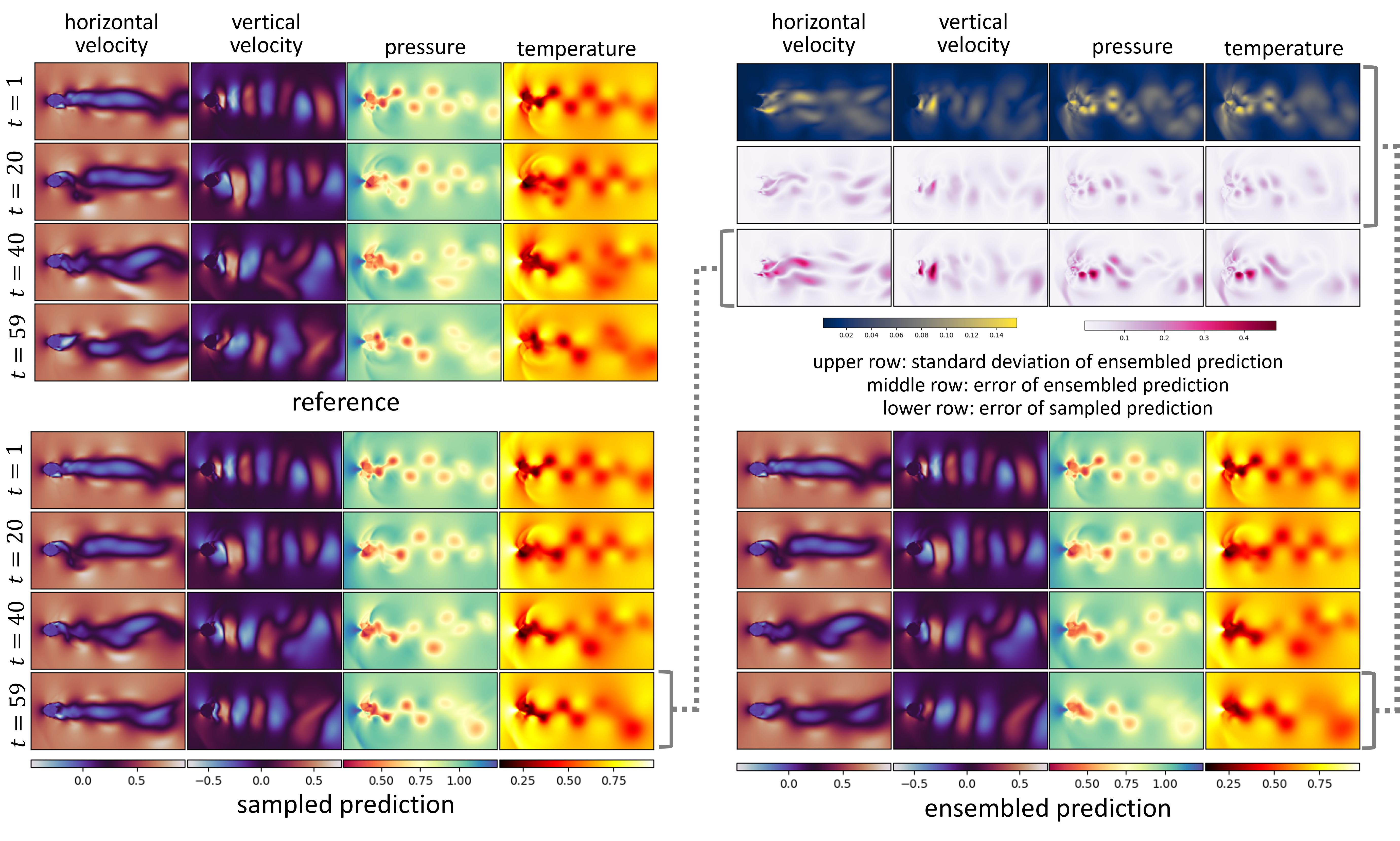}
        \caption{\small Visualization of data samples.}
        \label{fig:traj_2d_acdm}
        \end{subfigure}
    \end{minipage}
    \begin{minipage}{0.9\linewidth}
        \vspace{3mm}
        \begin{subfigure}{\linewidth}
        \includegraphics[width=1.0\linewidth]{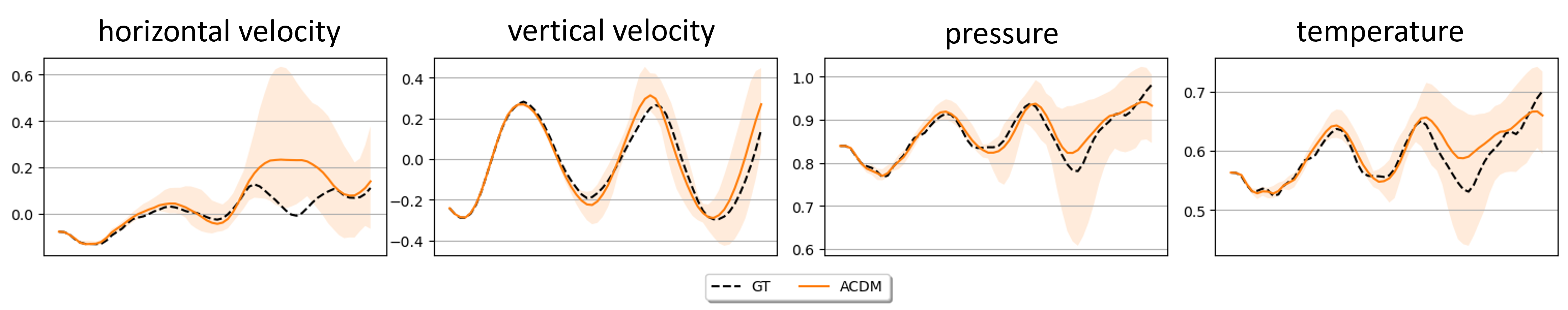}
        \caption{\small Visualization of prediction uncertainty at the center of the 2D domain.}
        \label{fig:traj_point_acdm}
        \end{subfigure}
    \end{minipage}
    \caption{A comparison of prediction samples and their ensemble from ACDM.}
\end{figure}

To evaluate the effect of ensembled prediction on ACDM, we used a checkpoint trained to simulate the transonic cylinder flow for Mach numbers $Ma\in [0.53,0.63]\cup [0.69, 0.90]$. At the inference stage, the model is tested on 6 simulation trajectories with Mach number ranging in $Ma\in [0.66,0.68]$. Each trajectory has a length of 60 timesteps and is simulated via rollouts. Similar to the experiment with PDE-Refiner, we sampled 16 rollouts from each inital condition, resulting in a data shape of $[B, T, D, C]$, where $B=6$ is the number of simulation trajectories, $T=59$ is the length of roll-out, $D=128\times 64$ is the spatial dimension, and $C=5$ is the number of data channels (horizontal velocity, vertical velocity, pressure, temperature, and Mach number). The sampled predictions and their ensemble (computed as the mean of all 16 samples) are evaluated with relative $\mathcal{L}^2$ prediction error as follows.
\begin{gather}
\label{eq:pred_error_acdm}
    e(t)=\frac{1}{B\cdot D\cdot C}\sum_{b=1}^{B}\sum_{d=1}^{D}\sum_{c=1}^{C}\frac{\|\Hat{u}(b,t,d,c)-u(b,t,d,c)\|_2}{\|u(b,t,d,c)\|_2},
\end{gather}
where $\Hat{u}$ is denotes a sampled prediction or the ensembled prediction. As shown in Fig. \ref{fig:pred_error_acdm}, the error of ensembled prediction is lower than any sampled prediction at the final timestep and for most parts of the rollouts. A visualization of data samples, point-wise prediction error and ensemble variance is provided in Fig. \ref{fig:traj_2d_acdm}. To visualize the uncertainty of model predictions, we show in Fig. \ref{fig:traj_point_acdm} a time sequence of ensembled prediction and its uncertain (defined as the range spanned by the minimum and the maximum predictions), evaluated at the center of the 2D domain. A quantitative comparison of prediction accuracy with and without ensemble is shown in the second row of Table \ref{tab:pred_error_and_residual_loss}, where ensembled prediction is shown to yield a lower prediction error than the mean prediction error of all 16 samples by the end of the simulation rollout.

\subsection{PI-DFS}
Our numerical experiments are conducted to examine the impact of ensembled prediction on the performance of the PI-DFS model, including impacts on the relative $\mathcal{L}^2$ prediction error, the residual loss and statistical metrics such as the kinetic energy spectrum and vorticity distribution. The task of reconstructing high-fidelity vorticity data from low-fidelity inputs was carried out on a test dataset of the shape $[B, T, D]$, where $B=4$ corresponds to four time sequences of turbulent flow vorticity in a 2D domain simulated by a numerical solver solving a Navier-Stokes equation, $T=318$ is the number of timesteps of each time sequence, and $D=256\times256$ specifies the size of the 2D domain. Similar to the experiments with PDE-Refiner and ACDM, given an input of low-fidelity vorticity data sampled from the test dataset, we generate 16 outputs from the PI-DFS model, and then compute their mean as the ensembled prediction of the high-fidelity vorticity data. The relative $\mathcal{L}^2$ loss and the relative residual loss (denoted as $e(t)$ and $r(t)$, respectively) are defined as follows.
\begin{gather*}
\label{eq:pde-refiner}
    e(b,t)=\frac{\|\Hat{\omega}(b,t)-\omega(b,t)\|_2}{\|\omega(b,t)\|_2}, \quad
    r(b,t)= \frac{1}{D}\sum_{d=1}^{D}\frac{|G(\Hat{\omega}_d)|^2}{\|\omega\|_2^2},
\end{gather*}
where $\omega$, $\Hat{\omega}$ denote the ground truth and the predicted vorticities of the turbulent flow, respectively. $G(\cdot)$ denotes a differential operator defined by the Navier-Stokes equation, \textit{e.g.}, $G(u,du/dt,\partial u/\partial \xi_1, \cdots, \partial^2 u/\partial \xi_1 \partial \xi_2, \cdots; \textit{Re})=0$, where $u$ is the velocity vector, $\xi_i$ is the spatial coordinate of the $i$-th direction, and $\textit{Re}$ is the Reynolds number. 

\begin{figure}[t]
    \centering
    \begin{minipage}{0.9\linewidth}
        \centering
        \includegraphics[width=1.0\linewidth]{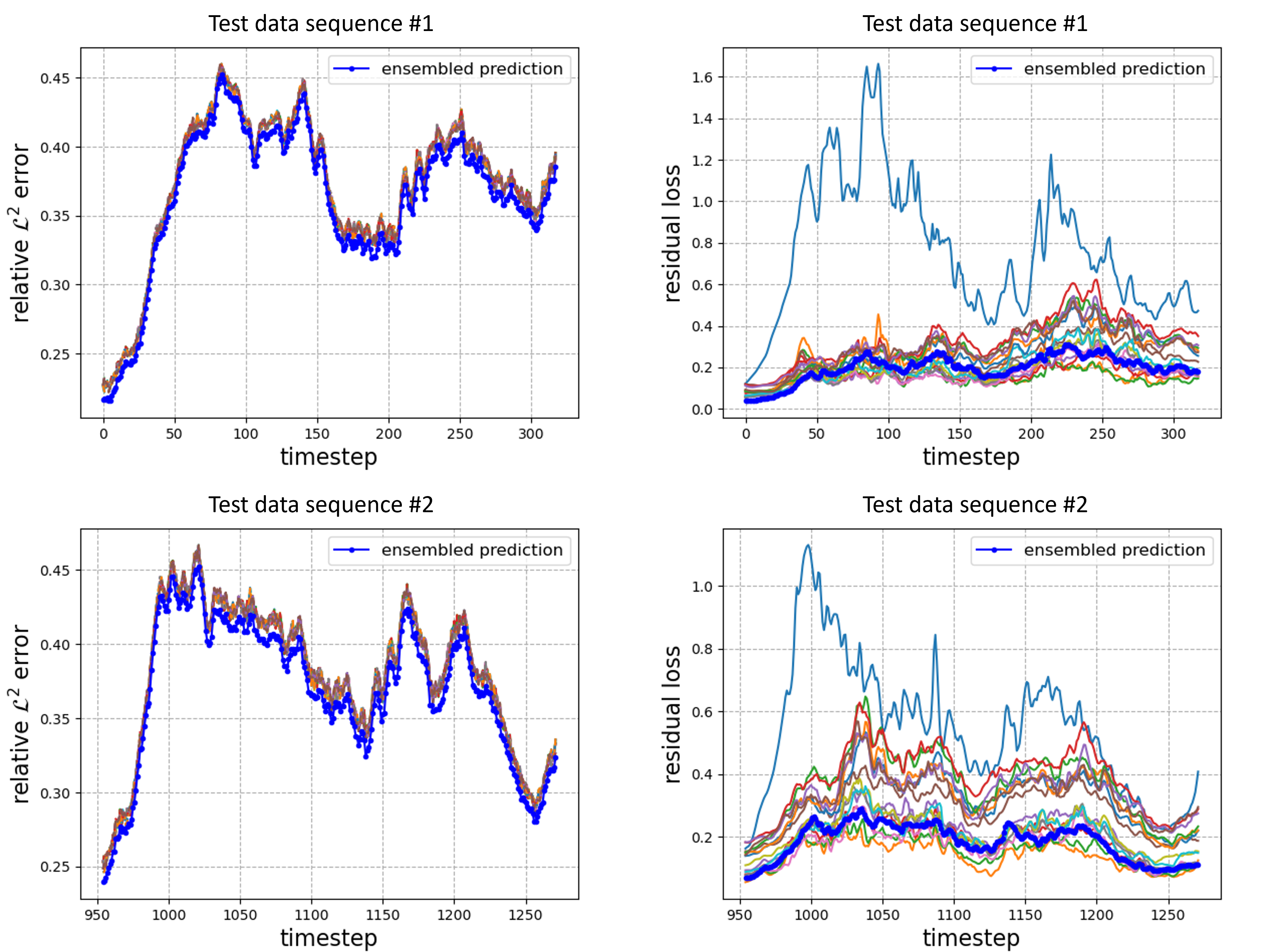}
        \caption{Prediction errors of model output samples and their ensemble on simulation trajectories from test data.}
        \label{fig:pred_error_dfsr}
    \end{minipage}
\end{figure}

\begin{figure}
    \centering
    \begin{subfigure}{0.45\textwidth}
        \centering
        \includegraphics[width=\linewidth]{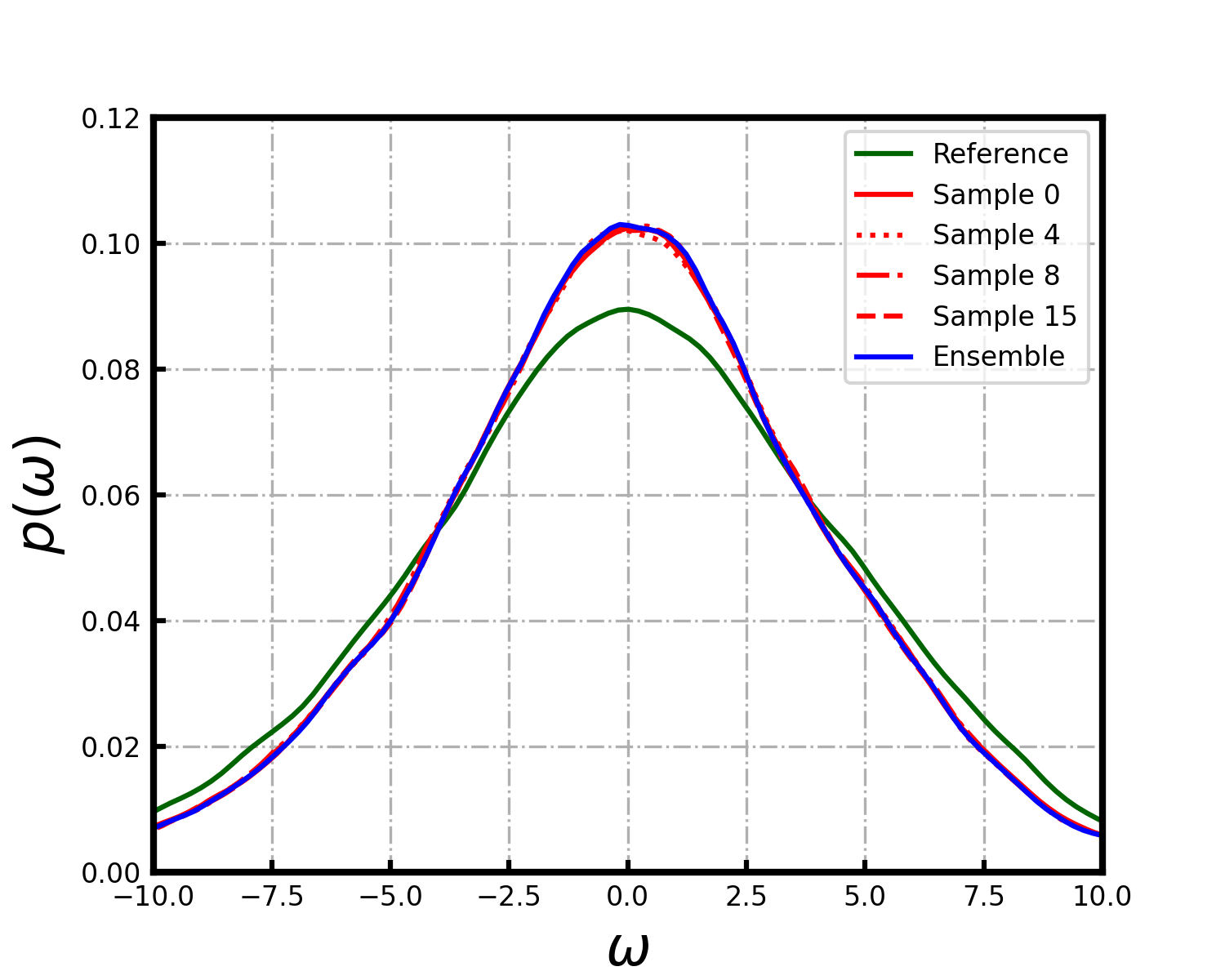}
        \caption{\small Vorticity distribution.}
        \label{fig:vorticity_distribution_dfsr}
    \end{subfigure}
    \begin{subfigure}{0.45\textwidth}
        \centering
        \includegraphics[width=\linewidth]{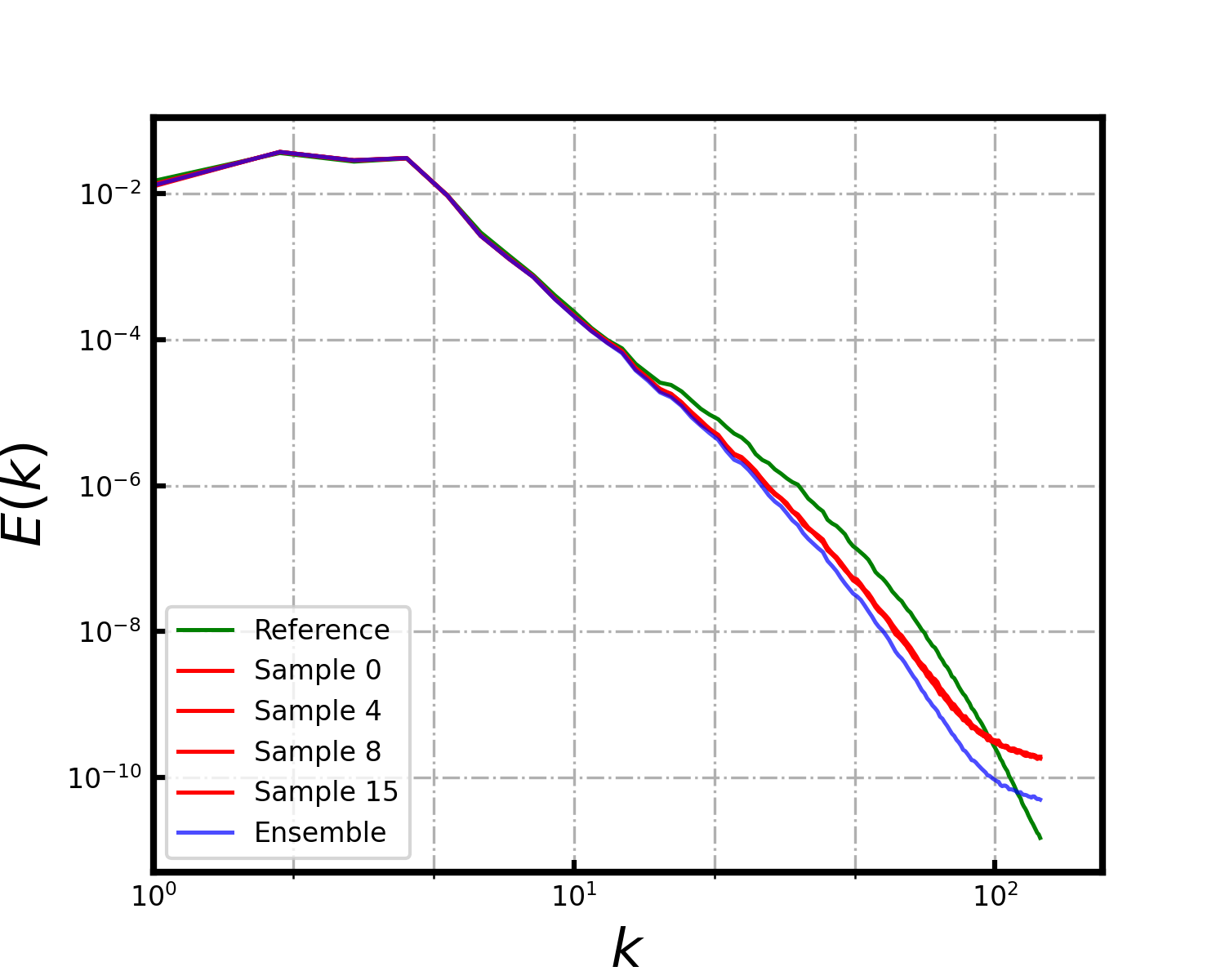}
        \caption{\small Kinetic energy spectrum.}
        \label{fig:energy_spectrum_dfsr}
    \end{subfigure}
    \caption{A comparison of model prediction samples and ensemble using statistical metrics.}\label{fig:samples_acdm}
    \label{fig:statistical_comparison_dfsr}    
\end{figure}

\begin{figure}
    \begin{minipage}{0.9\linewidth}
        \centering
        \vspace{3mm}
        \includegraphics[width=1.0\linewidth]{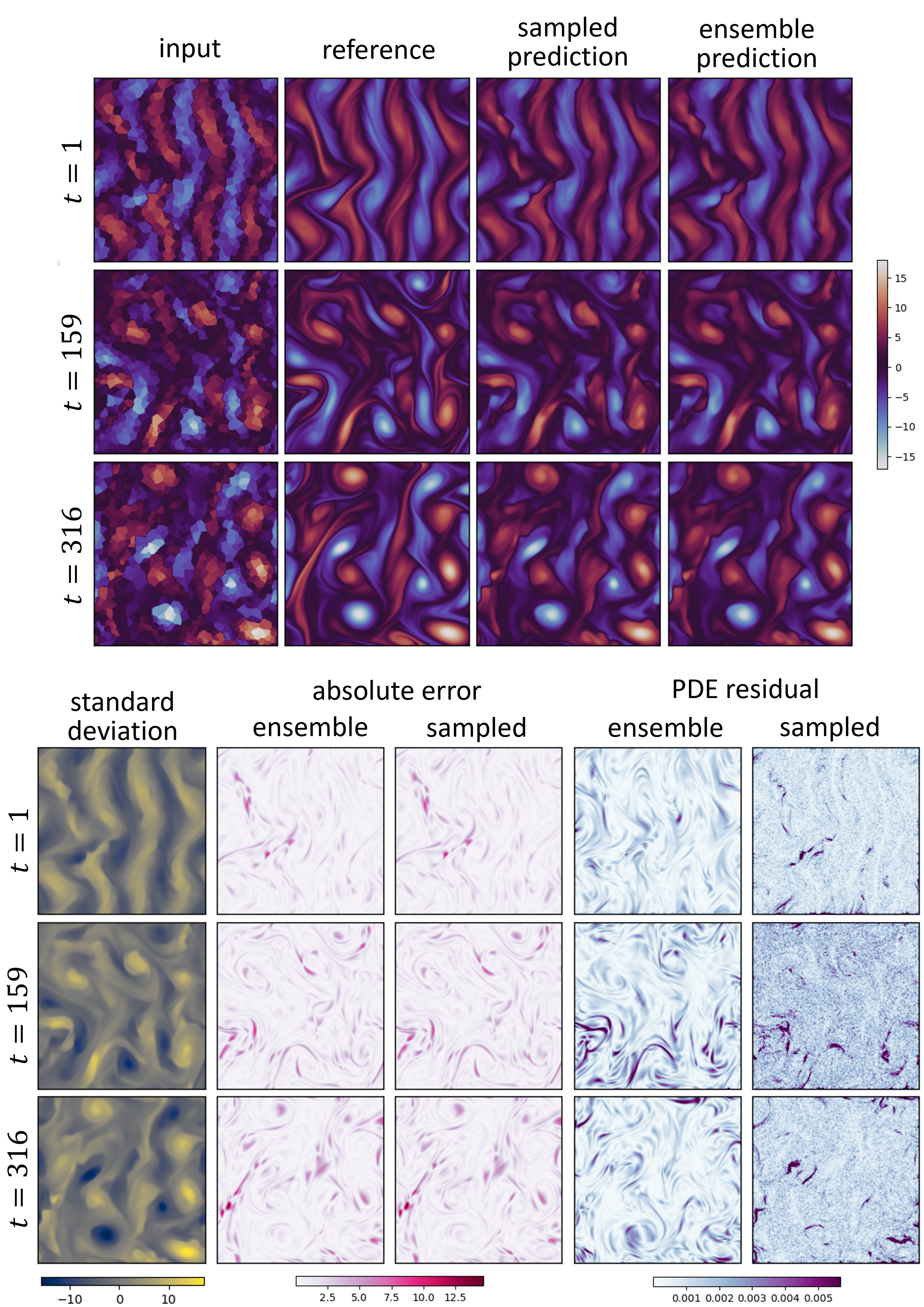}
        \caption{A comparison of prediction samples and their ensemble from PI-DFS.}
        \label{fig:sampled_visualization_dfsr}
    \end{minipage}
\end{figure}

Figure \ref{fig:pred_error_dfsr} shows the relative $\mathcal{L}^2$ loss and the relative residual loss of the PI-DFS model on two simulation trajectories sampled from the test dataset. In terms of $\mathcal{L}^2$ loss, the variation between different prediction samples is smaller than the error variation of PDE-Refiner and ACDM. The ensembled prediction yields a marginal reduction in the relative $\mathcal{L}^2$ loss. However, the residual losses from different prediction samples show a larger variation. Unlike the cases of the other two models, ensembled prediction with PI-DFS does not always have a residual loss lower than any sampled prediction. Nevertheless, ensemble still gains a significant decrease in residual loss from the mean residual loss of all prediction samples, as shown in the third row of Table \ref{tab:pred_error_and_residual_loss}. For a visualization of data samples used in the numerical experiment with PI-DFS, Figure \ref{fig:sampled_visualization_dfsr} shows the qualitative results of high-fidelity data reconstruction from a single prediction sample and from the ensemble of all prediction samples. Similar outcomes are observed in the visualization of the data samples (Columns 3 and 4, upper subfigure) and their point-wise absolute prediction errors (Columns 1 and 2, lower subfigure), whereas a more conspicuous difference can be observed in the point-wise absolute value of the PDE residual (Columns 4 and 5, lower subfigure). A visualization of point-wise variance of the ensembled prediction is also provided in Fig. \ref{fig:sampled_visualization_dfsr} (Column 1, lower subfigure), where it can be observed that a larger variance tends to appear in regions with higher vorticity. For a more comprehensive evaluation of ensembled prediction, we also compared the qualities of ensembled prediction and four samples prediction samples in a statistical sense using the metrics of kinetic energy spectrum and vorticity distribution, as shown in Fig. \ref{fig:statistical_comparison_dfsr}. In terms of vorticity distribution (Fig. \ref{fig:vorticity_distribution_dfsr}), ensembled prediction yields a highly similar result to these four samples. In terms of the kinetic energy spectrum each prediction, ensemble produces a marginally less accurate result than the four sampled predictions. As shown in Fig. \ref{fig:energy_spectrum_dfsr}, the kinetic energy spectrum of the ensembled prediction deviates further from the reference ground truth compared with that of any prediction sample on higher wave numbers. The high similarity between different prediction samples in Fig. \ref{fig:statistical_comparison_dfsr} indicates that the PI-DFS has a high consistency in the statistical accuracy of its prediction. The marginal reduction of accuracy in the energy spectrum of ensembled prediction is possibly a results losing some higher frequency spatial patterns due to the mean operation to obtain the ensembled prediction.

\end{appendix}

\end{document}